\documentclass[twocolumn,10pt]{simretractpreprint}
\usepackage{booktabs}
\usepackage{multirow}
\usepackage{placeins}
\newcommand{\method}{\textsc{SimRetract}}
\title[\method{}: Zero-Shot Sim-to-Real Robotic Assistance]{\method{}: Zero-Shot Sim-to-Real Robotic Assistance for Retractor Take-Over in Open Surgery}

\author[Wang et al.]{%
\textbf{Xucheng Wang}{, BS\textsuperscript{1,*}} \hfill{\small\mdseries\scshape davidx\_wang@hms.harvard.edu}\\
\textbf{Zhizhou Yang}{, MD\textsuperscript{1,2,*}} \hfill{\small\mdseries\scshape zyang17@mgh.harvard.edu}\\
\textbf{Sung Eun Kim}{, MD\textsuperscript{1}} \hfill{\small\mdseries\scshape sungeun\_kim2@hms.harvard.edu}\\
\textbf{Romain Hardy}{, MS\textsuperscript{1}} \hfill{\small\mdseries\scshape romain\_hardy@g.harvard.edu}\\
\textbf{Michael P. Sampson}{, BS\textsuperscript{2}} \hfill{\small\mdseries\scshape msampson1@mgb.org}\\
\textbf{Aakash Reddy}{, BS\textsuperscript{1,3}} \hfill{\small\mdseries\scshape aakash\_reddy@hms.harvard.edu}\\
\textbf{Ankit Pal}{, BS\textsuperscript{1}} \hfill{\small\mdseries\scshape ankit\_pal@fas.harvard.edu}\\
\textbf{Xiaoman Zhang}{, PhD\textsuperscript{1}} \hfill{\small\mdseries\scshape xiaoman\_zhang@hms.harvard.edu}\\
\textbf{Luyang Luo}{, PhD\textsuperscript{1,4}} \hfill{\small\mdseries\scshape luyang\_luo@hms.harvard.edu}\\
\textbf{Pranav Rajpurkar}{, PhD\textsuperscript{1}}\\
{\small\mdseries\itshape {\upshape\textsuperscript{1}}Department of Biomedical Informatics, Harvard Medical School}\\
{\small\mdseries\itshape {\upshape\textsuperscript{2}}Department of Surgery, Massachusetts General Hospital}\\
{\small\mdseries\itshape {\upshape\textsuperscript{3}}Weill Cornell Medical College}\\
{\small\mdseries\itshape {\upshape\textsuperscript{4}}Department of Biostatistics, Yale School of Public Health}\\
{\small\mdseries\itshape {\upshape\textsuperscript{*}}These authors contributed equally.}
}
\makeatletter
\let\ps@jmlrtps\ps@plain
\renewcommand*{\@jmlrenddoc}{}
\makeatother
\hypersetup{
  pdftitle={SimRetract: Zero-Shot Sim-to-Real Robotic Assistance for Retractor Take-Over in Open Surgery},
  pdfauthor={Xucheng Wang; Zhizhou Yang; Sung Eun Kim; Romain Hardy; Michael P. Sampson; Aakash Reddy; Ankit Pal; Xiaoman Zhang; Luyang Luo; Pranav Rajpurkar},
  pdfsubject={Preprint},
  pdfkeywords={sim-to-real, imitation learning, surgical robotics, retractor takeover},
  pdfdisplaydoctitle=true
}
\begin{document}
\maketitle

\begin{abstract}
We demonstrate the first zero-shot sim-to-real transfer for a surgeon-robot collaboration task in open surgery: retractor take-over. A robot arm takes a retractor from the surgeon's hand and preserves retraction at handoff, using a policy trained only on simulated demonstrations, without human demonstrations or real-world fine-tuning. Imitation learning from real demonstrations is limited by collection and labeling costs, and by performance degradation when the environment changes. Our framework, \method{}, generates simulated demonstrations designed for transfer: the policy observes masked instrument images and grasp geometry, learns to recover from off-nominal states, and acts with margin for execution drift. Deployed zero-shot, it grasps the retractor in 90\,\% of 50 surgeon-defined phantom presentations and preserves retraction at handoff in 78\,\%. Results on unseen retractors suggest potential for generalization to instruments with similar grasp geometry. Across ten policy configurations and 1\,200 simulated and real trials, we localize the sim-to-real gap to components: in simulation, grasp rates across ablations span 80--96\,\%, while on the real robot removing the instrument mask alone drops performance to 22\,\% grasped and 12\,\% held. Transfer-oriented design thus makes zero-shot sim-to-real transfer feasible for a surgical assistance task without human demonstrations.
Code is available \href{https://anonymous.4open.science/r/simretract-58D2}{here}.
\end{abstract}

\begin{keywords}
sim-to-real, imitation learning, vision-language-action models, open surgery, human-robot collaboration
\end{keywords}

\begin{figure*}[!t]
\centering
\includegraphics[width=0.80\textwidth]{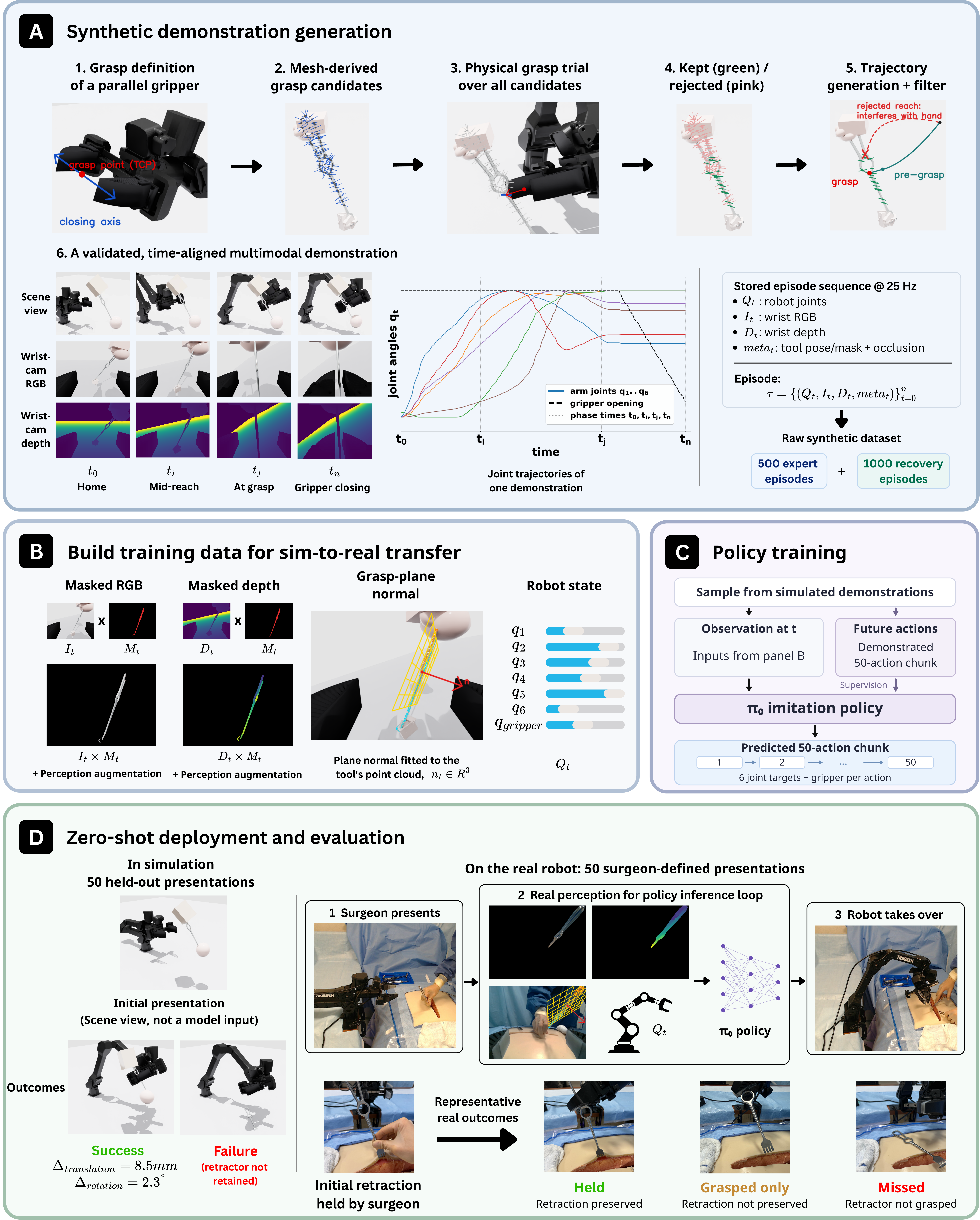}
\caption{\textbf{Overview of \method{}.} (A) A scripted expert generates complete approach and closure trajectories and retains physically validated demonstrations, including perturbation-recovery episodes. (B) Training inputs combine masked RGB, metric depth, robot joint state and grasp-plane normal, with perception augmentation. (C) Sampled observation--action pairs train $\pi_0$ to predict chunks of joint and gripper targets. (D) The same checkpoint is evaluated in simulation and deployed on the real robot without fine-tuning. Real outcomes at handoff are held, grasped only, or missed.}
\label{fig:pipeline}
\end{figure*}

\section{Introduction}
\label{sec:intro}

Open surgery relies on human assistants to provide anatomical exposure and maintain tissue tension through retraction. Although an experienced assistant can sometimes independently establish effective retraction, the primary surgeon commonly places the retractor to create the exposure and tension needed to operate, after which the assistant takes over and maintains the selected retraction. Here we focus on taking over a surgeon-positioned retractor, with the surgeon responsible for selecting the desired exposure. Because surgical personnel are scarce and unevenly distributed \citep{holmer2015global}, robotic assistance could help extend the capacity of existing teams \citep{kernbach2026humanoid}. The same scarcity limits expert time for collecting robot demonstrations, motivating simulation as a scalable source of training data. This handover requires perceiving an instrument partly occluded by the surgeon's hand and grasping it without disturbing retraction across unfamiliar environments.

We call this hand-off \emph{retractor take-over}. The surgeon establishes the desired tissue retraction and holds the instrument under load while the robot approaches and grasps the exposed shaft. A successful take-over requires more than grasping the retractor: the established retraction must be preserved at handoff, after the surgeon releases it.

The instrument is presented at varied positions and orientations and is partly occluded by the surgeon's hand. Only a limited portion of the shaft is available for grasping. A parallel gripper must align its pads with the shaft's flat faces and center the grasp site between them. Otherwise, closure can shift or rotate the instrument and alter the exposure even if the grasp succeeds. The robot must also approach without disturbing the loaded instrument and move smoothly and predictably beside the surgeon.

Imitation learning from real demonstrations is a natural route to such a skill \citep{black2024pi0,wang2026suture}, but prior work documents performance degradation under changes in background, lighting and viewpoint \citep{xie2024decomposing,pumacay2024colosseum}. Generalization to these shifts benefits from greater demonstration diversity \citep{lin2025datascaling}. Collecting this diversity across differently draped, lit and equipped surgical settings requires substantial expert time.

Simulation supports scalable data generation with precise labels and controlled variation \citep{tobin2017domain,makoviychuk2021isaacgym}. However, rendered observations differ from real ones, and surgical sim-to-real studies have largely focused on minimally invasive robotic platforms and surgical training tasks \citep{xu2021surrol,yu2024orbitsurgical,scheikl2023simtoreal,long2025embodied}. With the arm, gripper and wrist-camera configuration fixed, we ask whether a policy trained for retractor take-over on simulated demonstrations can perform the task in a previously unseen real environment without real task demonstrations or environment-specific policy fine-tuning. Wrist-camera sensing and robot joint state provide the policy input, without external cameras or a model of the room.

We propose \method{}, a framework for learning zero-shot retractor take-over from simulated demonstrations. The expert selects grasps and generates complete approach and closure trajectories, retaining demonstrations only after physical stability checks. Smooth, aligned approaches leave margin for execution drift, while perturbation-recovery adds corrective trajectories from perturbed poses. In both domains, the policy receives masked RGB-D, robot joint state and the grasp-plane normal. Training-time perception augmentation, including input dropout, mimics sensing imperfections.

We fine-tune $\pi_0$ \citep{black2024pi0} using a dataset of 1\,500 synthetic episodes and evaluate it in held-out simulation and on 50 surgeon-defined real presentations, with 1\,200 trials including ablations and unseen instruments. On the training instrument, the full policy achieves 90\,\% grasped and 78\,\% held at handoff in real-world trials. Masking has the largest effect on transfer, while the recovery-augmented dataset improves preservation of retraction.

\paragraph{Contributions.}
\begin{enumerate}
\item \textbf{The first zero-shot sim-to-real transfer for a surgeon-robot collaboration task in open surgery.} A policy trained on synthetic task demonstrations, with no real task demonstrations or real-world policy fine-tuning, grasps the retractor in 90\% of 50 surgeon-defined presentations and preserves the established exposure at handoff in 78\%.
\item \textbf{A framework for learning a surgical assistance task in simulation.} The framework generates and validates complete retractor take-over demonstrations, combines trajectories with execution margin and corrective recovery, and trains on task-relevant inputs with perception augmentation.
\item \textbf{A dissection of the sim-to-real gap for an open-surgery task.} Our comparisons cover ten policy configurations, 500 simulated and 700 real trials, including unseen instruments. In the main comparison, simulated grasp rates span 80--96\,\%, while real rates span 22--90\,\%; comparing both domains identifies which design choices support transfer.
\end{enumerate}

\section{Related Work}
\label{sec:related}

\paragraph{Robot assistance in surgery.}
Surgical autonomy research has concentrated on the minimally invasive setting and the da Vinci Research Kit: transformer policies from teleoperation \citep{kim2024srt} and hierarchical imitation learning that automates the clipping and cutting stages of ex-vivo cholecystectomy \citep{kim2025srth}; see \citet{schmidgall2025will}. Learning-based robotic assistance in \emph{open} surgery remains less explored than minimally invasive surgical automation. Existing work includes robotic scrub nurses, from gesture-driven systems \citep{jacob2013gestonurse} to VLA-based delivery \citep{li2024robonurse}, and an evaluation of imitation-learning policies on suture following \citep{wang2026suture}. Retractor take-over differs from delivery in that the robot must \emph{take over} an instrument held in a load-bearing pose while preserving the established retraction through the surgeon's release.

\paragraph{Generalization in imitation learning for robotics.}
In imitation learning a robot acquires a skill by reproducing demonstrated behavior from recorded observation--action pairs \citep{ross2011dagger}, with policies ranging from task-specific action decoders \citep{zhao2023act,chi2023diffusion} to pretrained vision--language--action (VLA) models such as $\pi_0$ \citep{black2024pi0}. Across architectures, policies degrade under changes of background, lighting and viewpoint \citep{xie2024decomposing,pumacay2024colosseum}, fail when their own errors carry them into states the demonstrations never covered \citep{ross2011dagger}. Generalization benefits from diverse training environments and objects \citep{lin2025datascaling}.

\paragraph{Simulation for scalable data and environment generalization.}
Simulation enables high-throughput data collection \citep{makoviychuk2021isaacgym} and scalable expansion of small human demonstration sets \citep{mandlekar2023mimicgen}. Domain randomization uses controlled variation to support transfer \citep{tobin2017domain,james2019simtoreal}. Videos of manual open surgery do not directly provide robot action targets, whereas instrumented research robots can record synchronized video and kinematics \citep{kim2024srt}. SurRoL \citep{xu2021surrol} and ORBIT-Surgical \citep{yu2024orbitsurgical} provide dVRK-compatible surgical learning environments. Surgical sim-to-real transfer has been demonstrated using SOFA \citep{scheikl2023simtoreal} and SurRoL \citep{long2025embodied}. \method{} takes a different strategy for the open field: rather than randomizing the whole scene, it removes environment-specific content from the policy input and corrupts the task-relevant perception signals that remain, keeping the policy end-to-end from masked pixels to joint targets.

\section{Method}
\label{sec:method}
Retractor take-over requires a grasp that preserves retraction and an approach without premature contact. Figure~\ref{fig:pipeline} outlines demonstration generation, policy inputs, training and deployment; Figure~\ref{fig:rollouts} shows representative sequences, with implementation details in Appendix~\ref{app:impl}.

\subsection{Synthetic demonstration generation}
\label{sec:sim}\label{sec:expert}\label{sec:recovery}
Our scripted expert combines grasp constraints, trajectories with execution margin and corrective recovery to generate complete take-over demonstrations.

\begin{figure*}[!t]
\centering
\includegraphics[width=0.78\textwidth]{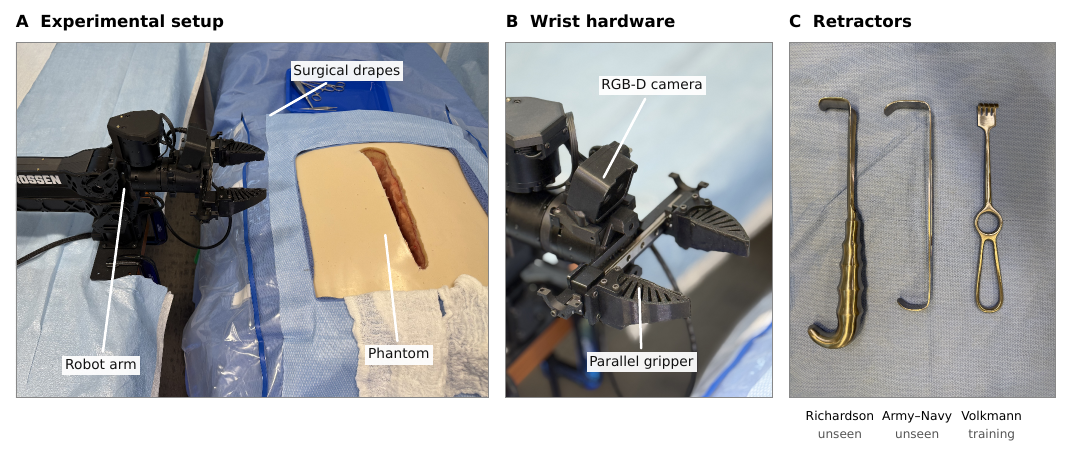}
\caption{Real experimental setup and instruments. (A) The arm beside the draped abdominal phantom. (B) The wrist-mounted RGB-D camera and parallel gripper. (C) Left to right: Richardson and Army--Navy, both unseen during training, and the Volkmann-type training retractor.}
\label{fig:robots}
\end{figure*}

\noindent\textbf{Scene.} Isaac Sim models the physical setup: a Trossen WidowX~AI arm, parallel gripper and wrist-mounted RealSense D405 (Figure~\ref{fig:robots}). A rigid Volkmann-type rake retractor is presented within a surgical-field-sized workspace in front of the arm (Figure~\ref{fig:training_sampling}; Appendix~\ref{app:simenv}). A randomized simulated hand occludes the shaft and holds it compliantly, allowing displacement on contact; a second occluder hides the hook end as tissue does. Retraction is represented by instrument pose, without deformable tissue mechanics. Low pad friction rejects grasps that rely on friction to compensate for poor alignment.

\noindent\textbf{Grasp.} The expert uses the simulator's exact instrument pose to select mesh-derived grasp candidates. The target centers the grasp site between the pads and aligns the closing axis with the shaft-plane normal, keeping pad faces parallel to the shaft faces to limit displacement during closure (Figure~\ref{fig:pipeline}A). Candidates must avoid approaching from below and clear the hand and tissue occluders.

\noindent\textbf{Trajectories with execution margin.} The policy predicts two-second \emph{action chunks} of joint targets, so abrupt direction changes between observations can cause overshoot or premature contact. \emph{Smoothing} combines a smooth reach curve with limited joint-target changes. A \emph{standoff} ends the curve short of the grasp for an aligned straight entry (Figure~\ref{fig:trajcmp}). Together these leave margin for execution drift. Inverse kinematics (IK) converts sampled gripper poses into arm joint targets. Only complete trajectories with undisturbed approaches and physically stable grasps are retained as demonstrations (Figure~\ref{fig:pipeline}A; Appendix~\ref{app:dataset}).

\noindent\textbf{Perturbation-recovery (PR).} To teach recovery from states outside nominal expert trajectories \citep{ross2011dagger,spencer2021feedback}, we interrupt an accepted demonstration at a uniformly sampled reach point, apply a small random pose perturbation and plan a fresh path to the grasp. Only the recovery segment is retained under the same physical criteria. These corrective data augment 500 expert demonstrations with 1\,000 PR episodes.

\subsection{Training data for precise sim-to-real grasping}
\label{sec:abstraction}
\noindent\textbf{Policy inputs.} The policy observation is $o=(M\odot I,\;M\odot\tilde D,\;[Q;n])$ (Figure~\ref{fig:pipeline}B), where $I$ is wrist RGB, $\tilde D$ metric depth, $M$ the binary instrument mask, $Q$ the six arm-joint positions and gripper position, and $n$ the shaft-plane unit normal. Here $\odot$ denotes pixelwise masking and $[Q;n]$ concatenation. Masking removes surrounding clutter, while depth and the normal provide geometry for precise grasp alignment. The normal is estimated by total-least-squares plane fitting with outlier rejection on the point cloud back-projected from masked depth pixels (Appendix~\ref{app:deployment}).

\noindent\textbf{Perception augmentation.} Training-time augmentation introduces random mask erosion, dilation, holes or leaks, depth speckle and edge artifacts, and plane-normal perturbations (Appendix~\ref{app:policy_training}). Close-range input dropout jointly blanks both images and the plane cue.

\subsection{Policy training}
\label{sec:training}
We fine-tune $\pi_0$ \citep{black2024pi0}, a pretrained vision--language backbone \citep{beyer2024paligemma} with an action expert that generates chunks by flow matching \citep{lipman2023flow}. The two masked images fill its camera slots, and $[Q;n]$ fills its state input. Each chunk contains $H=50$ targets at 25\,Hz, covering two seconds (Figure~\ref{fig:pipeline}C). The dataset is generated entirely in simulation: 1\,500 episodes and 167\,216 training frames after preprocessing (Appendix~\ref{app:dataset}).

\subsection{Zero-shot deployment}
\label{sec:deploy}
The wrist RGB-D camera supplies color and depth. SAM~3 \citep{carion2025sam3} tracks the retractor from a text prompt, and plane fitting on the masked depth point cloud gives $n$. With matched hardware, the same checkpoint predicts joint-target chunks without environment-specific policy fine-tuning (Figure~\ref{fig:pipeline}D). Inference requires no external camera, room model or room-registration step; camera preprocessing and controller details are in Appendix~\ref{app:deployment}.

\section{Experimental Protocol}
\label{sec:protocol}

\subsection{Ablation design}
\label{sec:ablations}
\method{} (full) is $\pi_0$ trained on 500 expert and 1\,000 perturbation-recovery (PR) episodes using all inputs and training augmentation. The seven variants in Table~\ref{tab:sim} test three design elements (\S\ref{sec:sim}--\ref{sec:abstraction}).

\noindent\textbf{Perception.} \emph{w/o mask} supplies unmasked RGB-D; \emph{w/o plane normal} removes the normal from the policy input. \emph{w/o dropout} disables joint image-and-plane input dropout.

\noindent\textbf{Demonstration coverage.} \emph{w/o recovery} retains only the 500 expert episodes. \emph{half data (250 + 500)} selects 250 unique grasp positions, retaining their 250 expert and 500 PR episodes. These compare corrective experience and coverage of grasp positions.

\noindent\textbf{Execution margin.} \emph{w/o smoothing} replaces the curved reach with two straight segments through a fixed 6\,cm pre-grasp and removes joint-rate limiting. \emph{w/o standoff} curves directly to the grasp without a separate aligned entry.

All variants use 15\,000 optimizer updates and 960\,000 training draws, including the data ablations (Appendix~\ref{app:policy_training}). Additional architectures are compared in Appendix~\ref{app:arch}.

\subsection{Simulation experiments}
\label{sec:simeval}
As a reference for task competence, we evaluate each policy on the same 50 held-out simulated presentations (Figure~\ref{fig:pipeline}D) using deployment code with the renderer's instrument mask replacing the real segmenter. A trial is \textbf{grasped} if the retractor drops by less than 20\,mm after the simulated operator releases it. For grasped trials, we measure post-release instrument-tip offset and axial twist relative to the presented pose. Twist is the axial component of the relative rotation vector; it complements tip offset by capturing changes in the retractor head's orientation. Scoring and termination rules are in Appendix~\ref{app:simscoring}.

\subsection{Real-robot experiments}
\label{sec:realeval}\label{sec:task}
Real trials use a Trossen WidowX~AI arm with a wrist RealSense D405 at a draped abdominal phantom under operating-room lighting (Figure~\ref{fig:robots}). Two attending surgeons each defined 25 presentations by establishing retraction at varied instrument positions and orientations, giving 50 presentations spanning plausible retraction directions. Each was photographed from two fixed viewpoints and re-created for every policy, with position and orientation checked against both views; outcomes are paired by presentation ID.

In each trial, the surgeon holds the retractor under load during the robot's approach and releases it after acquisition. Each trial permits one grasp attempt; robot motion stops at gripper closure, premature contact, or when the approach stalls or moves away from the retractor (Appendix~\ref{app:failures}). We also evaluate the full model and its no-plane ablation on two instruments unseen in simulation, Army--Navy and Richardson, with 50 paired presentations per instrument (25 per surgeon).\label{sec:generalization}

Simulation measures pose change through offset and twist; real trials assess its effect on tissue retraction and exposure. A trial is \textbf{missed} if the robot fails to secure the retractor or contacts the instrument, hand or phantom before grasping. Otherwise, it is \textbf{grasped} when the instrument remains in the gripper after release, and \textbf{held} when tissue retraction and exposure are preserved without repositioning at handoff; grasped includes both held and grasped-only trials (Figure~\ref{fig:pipeline}D). Miss subtypes are analyzed in \S\ref{sec:failures} and Appendix~\ref{app:failures}.

\section{Results}
\label{sec:results}

\subsection{A policy trained only in simulation takes over a surgeon-held retractor and preserves retraction on the real robot}
\begin{table*}[t]
\centering\footnotesize\setlength{\tabcolsep}{4pt}
\caption{Simulation and real-robot results (50 presentations each; outcomes in \S\ref{sec:simeval}--\ref{sec:task}). Variants are defined in \S\ref{sec:ablations}: w/o = without; half-data counts denote expert + recovery episodes. Offset and twist are post-release tip displacement and axial rotation-vector error over grasped simulated trials: mean $\pm$ SD [median]. Parentheses give descriptive real-minus-simulation grasp-rate differences (pp). Held denotes preserved retraction at handoff; paired comparisons are in Table~\ref{tab:paired}.}
\label{tab:sim}\label{tab:real}
\resizebox{\textwidth}{!}{%
\begin{tabular}{lcr@{\,$\pm$\,}r@{\,}lr@{\,$\pm$\,}r@{\,}l|cc}
\toprule
 & \multicolumn{7}{c|}{Simulation} & \multicolumn{2}{c}{Real robot}\\
Model & Grasped (\%) & \multicolumn{3}{c}{Offset (mm)} & \multicolumn{3}{c|}{Twist ($^\circ$)} & Grasped (\%) ($-$pp vs.\ sim) & Held (\%)\\
\midrule
\method{} (full) & 94 & 8.0 & 6.0 & [6.7] & 3.8 & 6.4 & [2.1] & \textbf{90} ($-$4) & \textbf{78}\\
w/o dropout & \textbf{96} & 9.9 & 7.5 & [8.1] & 5.9 & 11.5 & [2.5] & 88 ($-$8) & 68\\
w/o recovery & 80 & 15.6 & 21.0 & [7.1] & 6.0 & 8.8 & [4.8] & 78 ($-$2) & 34\\
half data (250 + 500) & 82 & 13.1 & 13.6 & [9.5] & 6.8 & 8.2 & [3.1] & 58 ($-$24) & 18\\
w/o plane normal & 84 & 20.0 & 21.2 & [12.0] & 10.9 & 18.3 & [3.9] & 76 ($-$8) & 58\\
w/o mask & 88 & 9.3 & 7.7 & [6.2] & 4.6 & 3.7 & [3.5] & 22 ($-$66) & 12\\
w/o smoothing & 86 & 8.8 & 7.7 & [7.3] & 4.5 & 4.3 & [3.8] & 54 ($-$32) & 42\\
w/o standoff & 84 & 10.2 & 14.8 & [5.5] & 2.6 & 1.9 & [1.9] & 68 ($-$16) & 36\\
\bottomrule
\end{tabular}}
\end{table*}

Trained only on simulated demonstrations, \method{} grasps the retractor in 90\,\% of 50 surgeon-defined real presentations and preserves retraction at handoff in 78\,\% (Table~\ref{tab:sim}; Figure~\ref{fig:gen}). The same checkpoint grasps 94\,\% of held-out simulated presentations, with a post-release tip offset of 8.0 $\pm$ 6.0\,mm and axial twist of 3.8 $\pm$ 6.4$^\circ$ over grasped trials. On the 45 grasped real trials, time to grasp is 10.08 $\pm$ 1.69\,s (median 10.10\,s; Appendix~\ref{app:time}).

\subsection{Simulated success does not fully predict real success; masking has the largest effect on transfer}
Across the eight configurations in Table~\ref{tab:sim}, simulated grasp rates span 80--96\,\%, while real rates span 22--90\,\%. Removing the instrument mask reduces grasp success by six percentage points in simulation (94\,\% to 88\,\%), compared with 68 points on the real robot (90\,\% to 22\,\%; Holm-adjusted $p<0.001$ for the real comparison). The sim-to-real grasp gap consequently expands from four percentage points to 66.

Motion design also has larger effects after transfer, although the contrast is smaller than for masking. Removing smoothing costs eight percentage points of grasp success in simulation and 36 on the real robot; removing standoff costs ten and 22 points, respectively.

\subsection{Corrective demonstrations and execution margin turn a grasp into preserved retraction}
\label{sec:failures}
Recovery and standoff produce larger improvements in held success than in grasped success. Recovery-augmented training raises grasp success from 78\,\% to 90\,\% and held success from 34\,\% to 78\,\%. Including standoff raises grasp success from 68\,\% to 90\,\% and held success from 36\,\% to 78\,\% (both held comparisons: Holm-adjusted $p<0.001$; Table~\ref{tab:paired}).

Among presentations grasped by both the full model and the no-recovery variant, retraction was preserved only by the full model in 17 cases and only by the no-recovery variant in one. For the full model versus the no-standoff variant, the corresponding counts were 13 and two.

\begin{figure*}[t]
\centering
\includegraphics[width=0.86\textwidth]{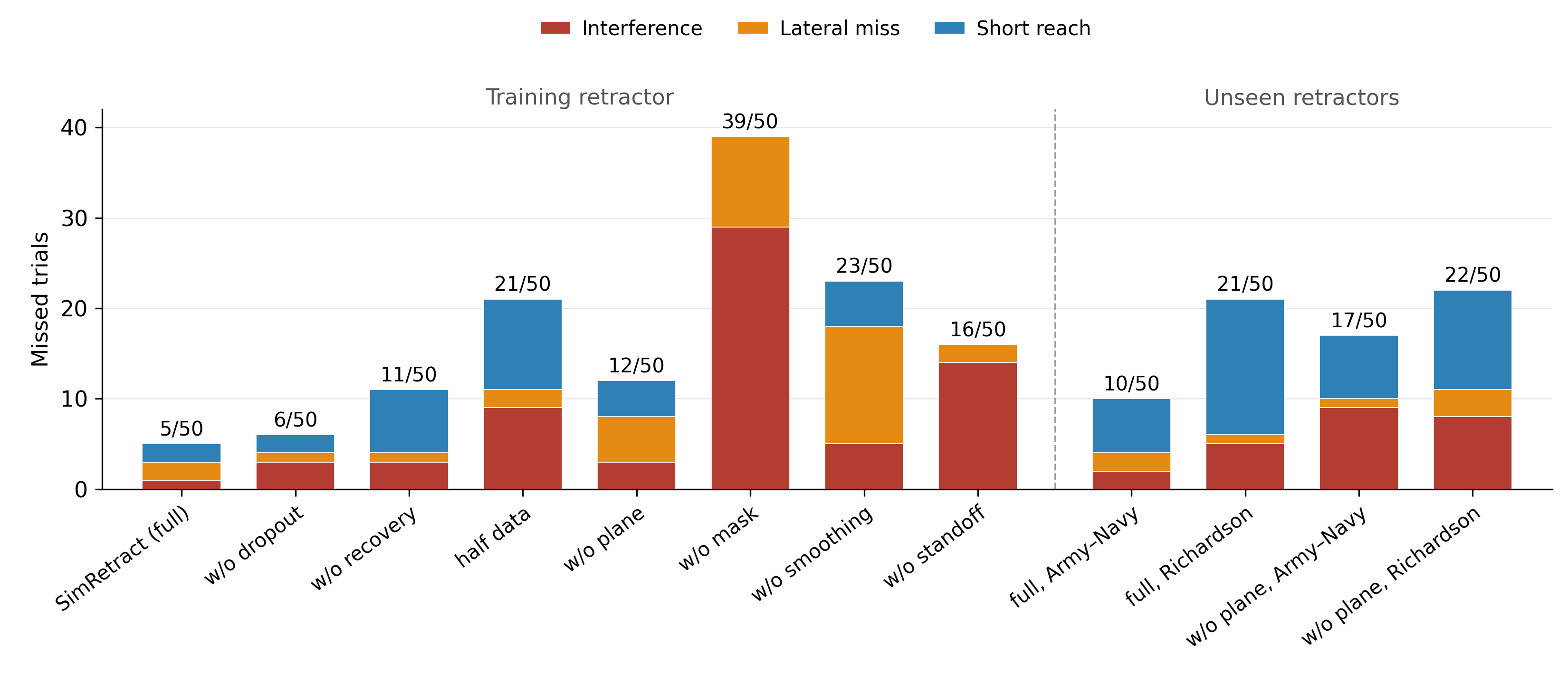}
\caption{Failure types in real trials (50 per model; bar labels: missed/evaluated). Interference is pooled; contact locations are in Table~\ref{tab:interference}. Grasped-only trials are excluded.}
\label{fig:failures}
\end{figure*}

Removing smoothing also lowers held success, from 78\,\% to 42\,\% (Holm-adjusted $p=0.006$). The ablations exhibit distinct approach failures: short reaches account for 7 of 11 misses without recovery, lateral misses for 13 of 23 without smoothing, and premature interference for 14 of 16 without standoff (Figure~\ref{fig:failures}; contact locations in Table~\ref{tab:interference}).

Despite having fewer episodes, the 500-episode nominal-only model, covering 500 grasp positions, achieves higher grasp success than the 750-episode half-data model covering 250 positions: 78\,\% versus 58\,\% (unadjusted $p=0.041$).

\subsection{Transfer extends to unseen retractors, with higher success for similar grasp geometry}
\label{sec:genresults}
\begin{figure}[t]
\centering
\includegraphics[width=\columnwidth]{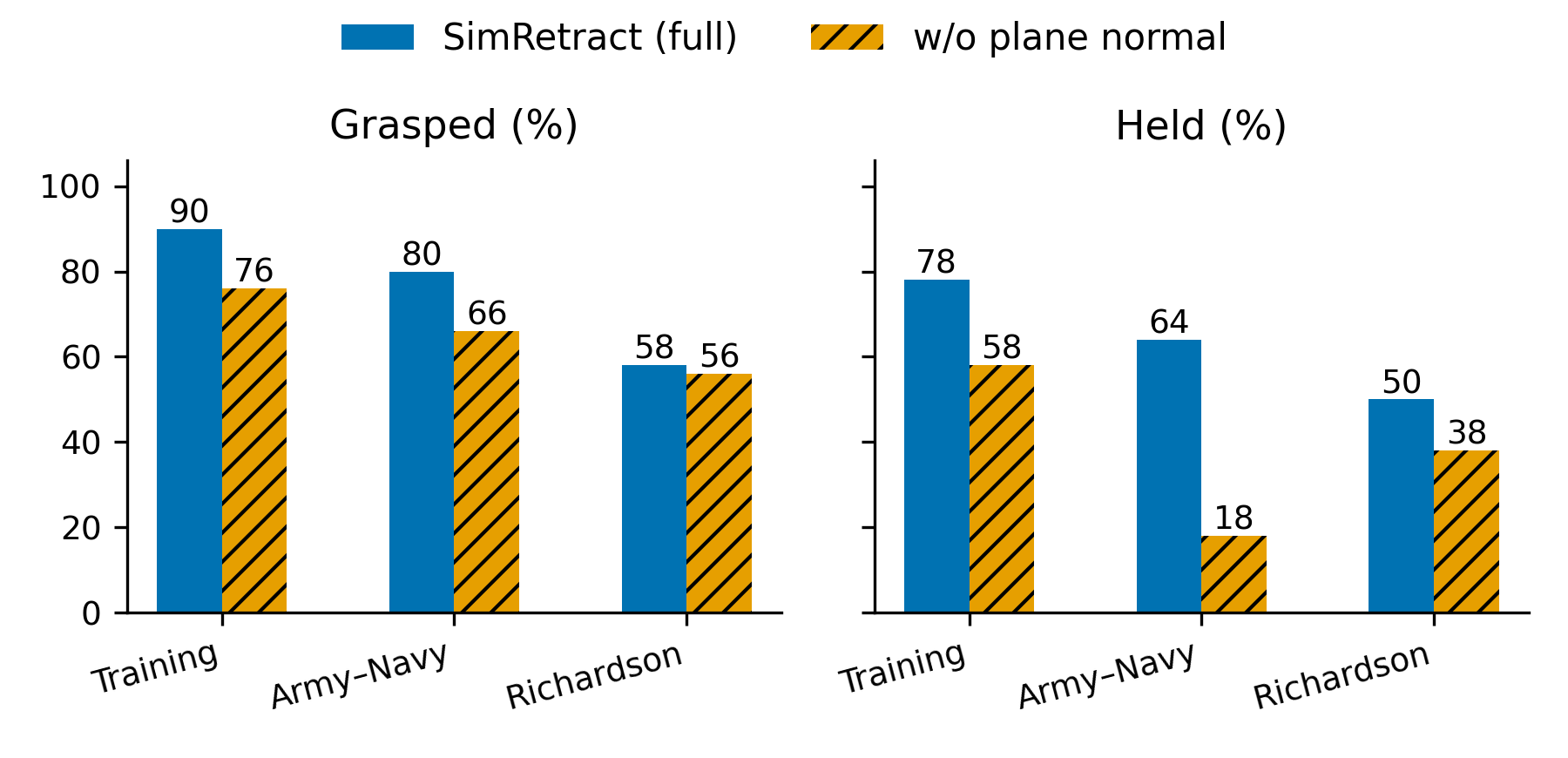}
\caption{Generalization to unseen retractors (50 presentations each): grasped and held rates at handoff for the full model and the no-plane ablation.}
\label{fig:gen}
\end{figure}

On the unseen Army--Navy, which shares the training instrument's flat grasp surface, the full model achieves 80\,\% grasped and 64\,\% held. On Richardson's cylindrical handle, it achieves 58\,\% grasped and 50\,\% held (Figure~\ref{fig:gen}). Removing the plane cue lowers Army--Navy performance to 66\,\% grasped and 18\,\% held. Richardson grasp rates change little with and without the cue (58\,\% versus 56\,\%), while held rates are 50\,\% versus 38\,\%. Short reaches dominate Richardson misses both with and without the plane cue, accounting for 15 of 21 and 11 of 22 misses, respectively.

\FloatBarrier

\section{Discussion}
\label{sec:discussion}
\method{} supports a form of surgical assistance in which the surgeon establishes the desired retraction and exposure, and the robot takes over the hold. Learning this behavior from simulated demonstrations offers a route to developing assistance skills without collecting demonstrations from surgeons. Its significance lies in preserving a surgeon-selected state during handoff, beyond acquiring and retaining an instrument.

\paragraph{Task-relevant perception is central to zero-shot transfer.}
Masking reduces dependence on surrounding appearance while retaining the instrument and its grasp geometry. Its disproportionate benefit on the real robot supports designing observations around information that remains useful across simulation and deployment. The mask ablation identifies this input design as decisive for transfer in the tested framework, while the motion ablations motivate a complementary focus on accommodating deviations during execution.

\paragraph{Demonstrations need margin for execution drift.}
The paired recovery and standoff comparisons link demonstration design to preserving retraction even when acquisition succeeds in both conditions. Perception errors and differences in controller response can push real rollouts away from the trajectories seen during training. Recovery episodes provide experience correcting these deviations; smoothing combines curved reaches and joint-rate limiting, while standoff leaves room to align before contact. These designs allow correction before misalignment at closure disrupts the established retraction. The nominal-only and half-data comparison further suggests that broader grasp-position coverage can be more valuable than additional episodes around fewer grasps.

\paragraph{Shared grasp geometry provides a reusable basis for transfer.}
The Army--Navy result supports the grasp-plane normal as a reusable cue when instruments share a local grasp surface, despite differences in their overall form. Richardson illustrates how changes in appearance can influence the use of these geometric inputs: its short reaches suggest that the policy may interpret a wider mask as closer and underuse metric depth. Preserving meaningful grasp cues while strengthening the use of depth could extend transfer across instrument shapes.

\paragraph{Limitations and future work.}
We designed the phantom evaluation to test robustness across a wide range of plausible retractor positions and orientations. The full model failed to preserve retraction at six corresponding presentations across all three retractors (Appendix~\ref{app:paired}), identifying shared difficult configurations. Studies with biological tissue or animal models should next assess preservation of exposure during active surgeon--robot collaboration.

Collision avoidance is currently implicit in the learned grasp and approach behavior: demonstrations favor exposed-shaft grasps, smooth motion and standoff, while deployment lacks explicit online collision monitoring. Interference with the gripper's long rail and wrist camera also motivates a more compact end-effector. Future work should combine contact sensing and active collision monitoring with safe stopping, recovery and return of control when the surgeon, another instrument or the tissue moves.

\section{Conclusion}
\method{} demonstrates zero-shot sim-to-real retractor take-over on a general-purpose robot arm, preserving surgeon-established retraction at handoff with a policy trained only on simulated demonstrations and no human demonstrations or real-world policy fine-tuning. The findings identify task-relevant perception, corrective demonstrations and execution margin as key elements of transfer, with shared grasp geometry supporting reuse across instruments. These results support simulation-based learning of surgical assistance and motivate further evaluation with biological tissue and active surgeon--robot interaction.

\bibliography{ref}

\appendix
\clearpage\onecolumn\clearpage
\setcounter{figure}{0}
\setcounter{table}{0}
\renewcommand{\thefigure}{S\arabic{figure}}
\renewcommand{\thetable}{S\arabic{table}}
\ifcsname theHfigure\endcsname
  \renewcommand{\theHfigure}{supp.\arabic{figure}}
\fi
\ifcsname theHtable\endcsname
  \renewcommand{\theHtable}{supp.\arabic{table}}
\fi
\section{Representative rollouts}
\label{app:visuals}\label{app:rollouts}
\begin{figure}[!ht]
\centering
\includegraphics[width=0.84\textwidth]{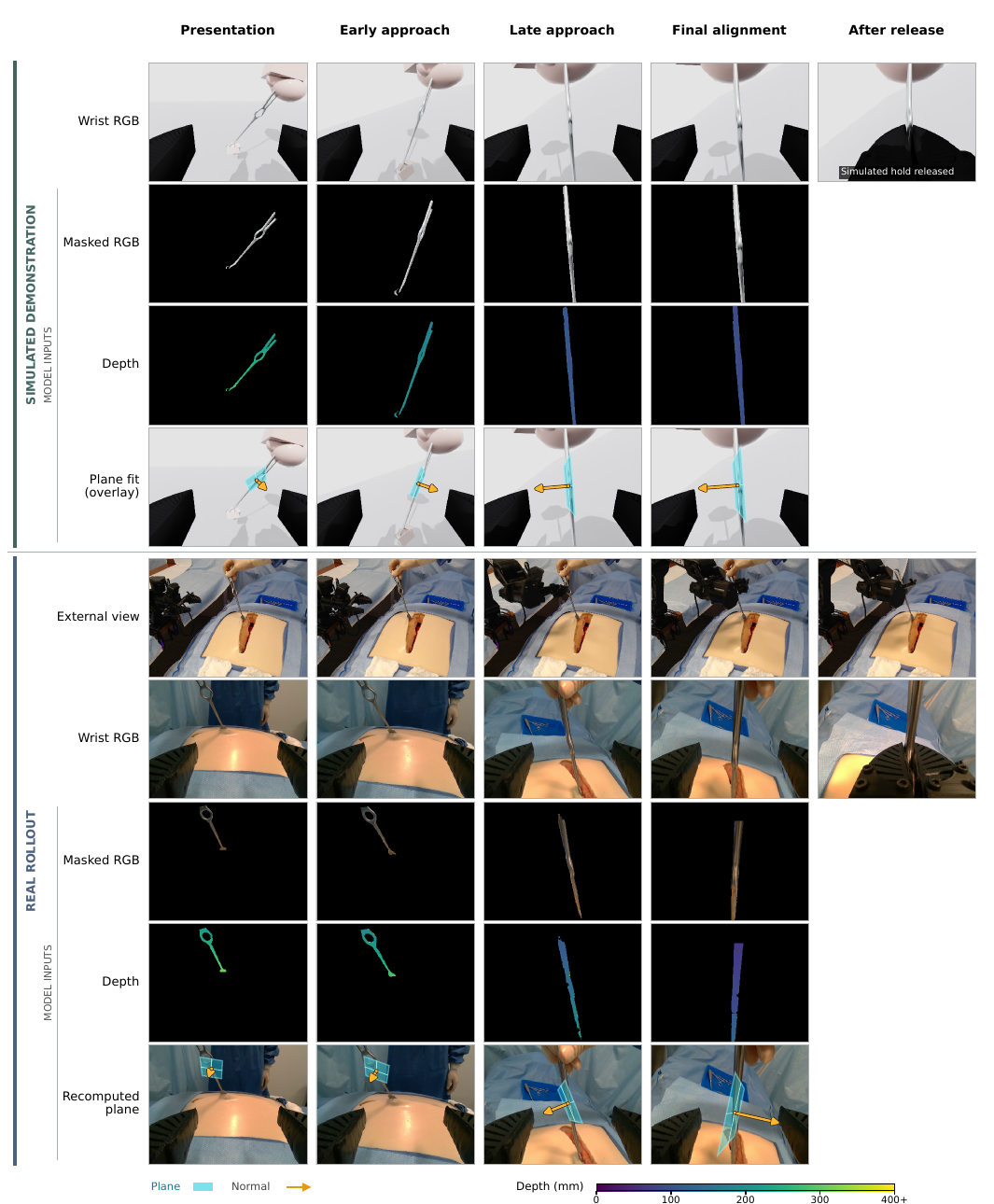}
\caption{Simulated demonstration and real rollout across five stages. Inputs include masked RGB, metric depth, plane normal and joint state. Plane overlays show the fitted plane (cyan) and normal (yellow). Simulation inputs precede augmentation. Real RGB-D matches recorded queries. Plane overlay illustrations are recomputed from recorded RGB-D. Input panels end before release; the simulated hand proxy remains visible.}
\label{fig:rollouts}
\end{figure}

Figure~\ref{fig:rollouts} shows a simulated demonstration and real take-over.

\clearpage\twocolumn

\section{Implementation details}
\label{app:impl}

\subsection{Simulation environment}
\label{app:simenv}
\paragraph{Robot and scene.}
We use Isaac Sim~5.1 / Isaac Lab~2.3 \citep{mittal2025isaaclab} with the manufacturer's Trossen WidowX~AI model and the flat-pad fingers used on the real arm. The wrist camera renders at $640\times480$ with the real D405 intrinsics ($58^\circ$ vertical field of view); physics runs at 500\,Hz, observations and actions at 25\,Hz. The instrument is a rigid mesh of a 21.6\,cm Volkmann-type rake retractor, with a pronged hook and flat ring handle, rendered as stainless steel. Its mass is randomized over 60--160\,g and pad friction is deliberately low to reject grasps that rely on high friction. One seed fixes the presented pose, mass, hand occluder and lighting, including dome and key intensity, color temperature, key direction and softness, and up to three fill lights.

\begin{figure}[!ht]
\centering
\includegraphics[width=0.80\columnwidth,trim=220bp 200bp 320bp 260bp,clip]{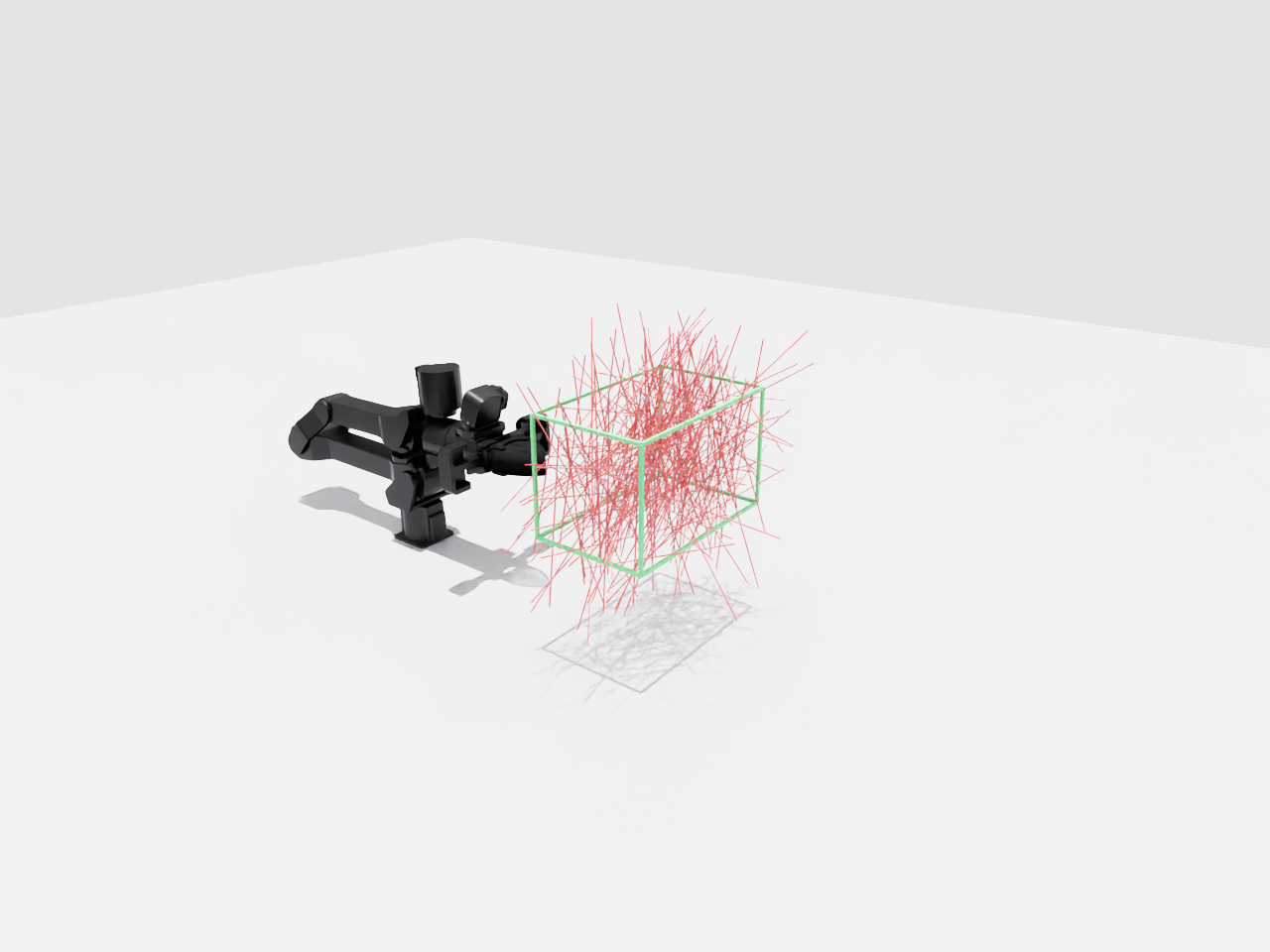}
\caption{Instrument-pose sampling domain relative to the robot. The green box bounds the proposed instrument centers; red lines show example shaft positions and orientations. The shaft can extend outside the center-sampling box.}
\label{fig:training_sampling}
\end{figure}

\paragraph{Presentation region.}
Instrument centers are sampled uniformly in a $15\times24\times15$\,cm box representing a draped surgical field in front of the robot: 35--50\,cm forward of the base, $\pm12$\,cm laterally and 15--30\,cm above the base-frame origin. These limits keep the presentations within the arm's reach and also bound the real phantom presentations. The box bounds the instrument's center; the 21.6\,cm retractor can extend about 11\,cm beyond it, giving an approximate outer envelope of $37\times46\times37$\,cm. Tilt from vertical is sampled uniformly up to $80^\circ$, with uniform azimuth and axial spin (Figure~\ref{fig:training_sampling}).

\paragraph{Operator hold and occlusion.}
The simulated hold is a spring--damper on the presented pose, with yield-capped torque about the two tilt axes and a weaker cap about shaft spin. Under the instrument's weight it sags by a fraction of a millimeter; a firm push at the tip (about 5\,N) rotates it by tens of degrees. Thus a misaligned approach moves the instrument rather than being absorbed by a rigid attachment. A randomized fist or pinch occludes the shaft below the grasp region. A second visual occluder, a 1--3\,cm ellipsoid or multi-primitive blob in tissue and drape tones, covers the hook end to represent the loaded tip being visually occluded by tissue. Neither enters the instrument mask. Both receive a 6\,mm safety margin in clearance tests; robot entry into either region aborts demonstration generation. The scene models instrument pose and hand compliance, without deformable tissue mechanics.
\FloatBarrier

\subsection{Grasp generation}
\label{app:grasps}
A bank of 432 candidate grasps is computed once from the instrument mesh, using face-on clamps on near-parallel opposing faces and antipodal pairs on non-flat parts. For each presentation, only the exposed instrument region included in the mask is treated as graspable. Grasps and approach paths avoid interference with the hand and tissue regions excluded from the mask. The closing axis is aligned with the shaft-plane normal, making the pad faces parallel to the shaft faces. The approach axis lies approximately in that plane; candidates vary the remaining rotation about the closing axis. A target grasp height is sampled uniformly over $\pm5$\,cm along the shaft; candidates are tried in order of distance from that height. For each candidate, the expert generates and tests a complete approach and closure trajectory; the first passing the physical checks in Appendix~\ref{app:dataset} is retained as a demonstration.

Each of the 500 poses used for nominal demonstrations has 432 mesh-derived grasp candidates, giving 216\,000 candidates in total. Only 5.89\,\% passed physical grasp trials, highlighting the need for execution-level validation of geometry-based grasp candidates.

\subsection{Trajectory generation}
\label{app:traj}
Each selected grasp defines a target gripper position $p_g$ and orientation $R_g$. The expert first constructs a path of gripper poses from the current pose to this target, then converts them into arm joint targets using inverse kinematics (IK).

\paragraph{Gripper path.}
The grasp orientation $R_g$ defines the approach direction $a$. The full expert places the pre-grasp 3\,cm before the grasp, $p_{\mathrm{pre}}=p_g-0.03a$, and a control point 6\,cm farther back, $p_c=p_{\mathrm{pre}}-0.06a$ (coordinates in meters). From the starting position $p_0$, it follows the quadratic B\'ezier curve
\begin{equation}
B(t)=(1-t)^2p_0+2(1-t)t\,p_c+t^2p_{\mathrm{pre}},
\label{eq:bezier}
\end{equation}
for $t\in[0,1]$, then enters straight along $a$ for the final 3\,cm. The control point aligns the curve's final tangent with this entry. Gripper orientation interpolates from its initial orientation to $R_g$ over the first 70\,\% of the curve and remains fixed during entry. The path is sampled at a nominal gripper speed of about 0.11\,m/s.

\paragraph{IK and joint-rate limiting.}
At each path sample, a damped least-squares IK controller \citep{wampler1986dls} uses the arm's current state and the error to the desired gripper position and orientation to compute six arm joint-position targets. Smooth gripper motion does not guarantee smooth motion at every joint: the mapping depends on the arm's configuration, so even a nearly constant gripper speed can require abrupt rotation of an individual joint. To limit these spikes, large changes between successive joint targets are subdivided into intermediate commands, limiting each commanded joint-position increment to $2.5^\circ$. This slows the arm locally where needed, leaving margin for execution drift. At the grasp, the gripper closes until both pads report contact, then presses. The resulting arm and gripper commands form the demonstration actions; the learned policy subsequently predicts joint-target chunks directly.

\begin{figure}[!ht]
\centering
\includegraphics[width=\columnwidth]{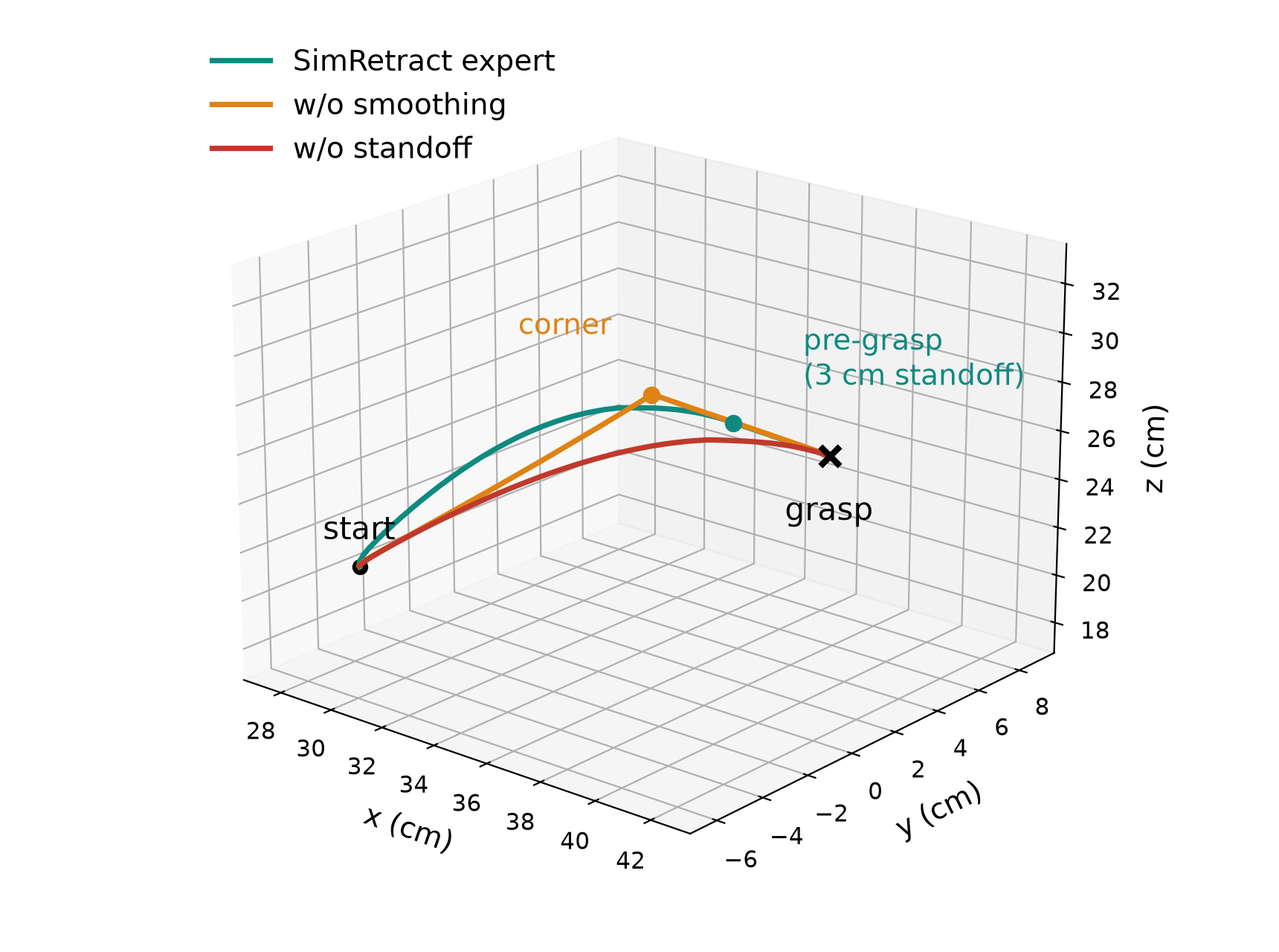}
\caption{TCP paths reconstructed from joint states for the same presentation and grasp. \method{} (teal): smooth curve, then a 3\,cm aligned straight entry; \emph{w/o smoothing} (orange): straight reach and aligned entry; \emph{w/o standoff} (red): curve directly to the grasp.}
\label{fig:trajcmp}
\end{figure}

\paragraph{Trajectory variants.}
The \emph{w/o smoothing} expert uses two straight segments: a reach to a pre-grasp 6\,cm before the grasp, followed by an aligned straight entry. It uses neither a B\'ezier curve nor joint-rate limiting. The \emph{w/o standoff} expert follows one B\'ezier curve directly to the grasp, using its candidate pre-grasp as the control point, with no separate straight entry. Both use IK to convert their gripper paths into joint targets. Figure~\ref{fig:trajcmp} illustrates the three constructions.

\FloatBarrier

\subsection{Training-set construction}
\label{app:dataset}
\paragraph{Physical filtering.}
During demonstration generation, an approach is rejected if it moves the instrument by more than 3\,mm or $3^\circ$ before closure, balancing grasp precision with generation efficiency. After closure and release of the simulated hold, an episode is retained if translation from the presented pose is below 10\,mm and rotation is below $10^\circ$. The 500 unperturbed (nominal) expert trajectories are balanced across four 2.5\,cm grasp-height bands spanning $\pm5$\,cm along the shaft, with 125 episodes per band.

\paragraph{Perturbation-recovery episodes.}
Four recovery draws are proposed per nominal seed. Each perturbs the robot 25--75\,\% of the way along the approach trajectory. The lower cutoff targets recovery after the initial action chunk, which always starts from home; the upper cutoff reduces the risk of random perturbations driving the gripper into the retractor near contact. The arm is offset by 3--8\,cm (scaled down for later perturbations), rotated by 8--25$^\circ$, and its gripper part-closed with 40\,\% probability. A fresh trajectory is generated from the perturbed state and kept under the same physical criteria; only post-perturbation frames are recorded. The training subset uses interruptions in the first half of the reach. Its 1\,000 PR episodes use the same pool of 500 nominal presentations.

\paragraph{Final dataset and preprocessing.}
The curated dataset contains 500 nominal expert demonstrations and 1\,000 PR episodes. Episodes are trimmed one second after the gripper reaches approximately its final closed width, then resampled by retaining frames when joint motion exceeds a small quantum, reducing repeated stationary observations. The resulting 167\,216 training frames comprise 65\,784 nominal and 101\,432 recovery frames. Training samples the two components equally, irrespective of their episode or frame counts. Data generation uses one RTX~5090 with up to eight simulator instances in parallel.

\subsection{Policy training}
\label{app:policy_training}
\paragraph{Policy hyperparameters.}
All reported models use 15\,000 optimizer updates with an effective batch size of 64, totaling 960\,000 sampled observation--action-chunk pairs, including repeated samples. The data ablations change the episode pool while preserving this training budget (\S\ref{sec:ablations}); dataset composition and sampling weights are given in Appendix~\ref{app:dataset}. We report the final checkpoint without model selection. The main policy uses $\pi_0$; SmolVLA \citep{shukor2025smolvla} is fully fine-tuned with the same optimizer settings and training budget (Table~\ref{tab:hyper}). During training and deployment, both VLA policies receive the fixed instruction ``take over the tool from the operator and hold its pose''.

\begin{table}[h]
\centering\footnotesize
\caption{Shared policy-training hyperparameters. Sample counts include repeated draws from each model's training pool.}
\label{tab:hyper}
\begin{tabular}{@{}ll@{}}
\toprule
Optimizer & AdamW\\
Peak learning rate & $2.5\times10^{-5}$\\
Schedule & 1\,k warm-up, cosine decay\\
Effective batch size & 64 ($16\times4$ accumulation)\\
Optimizer updates & 15\,000\\
Sampled training pairs & 960\,000\\
Action chunk & 50 targets (2\,s at 25\,Hz)\\
\bottomrule
\end{tabular}
\end{table}

On one H200, $\pi_0$ training takes 1.65\,s per optimizer update (about 7\,h for 15\,000 updates), and SmolVLA 0.92\,s. ACT with input dropout takes approximately 0.26\,s per update.

\paragraph{Training augmentation.}
Augmentation is sampled online during training: instrument masks and depth images are corrupted, and the estimated grasp-plane normal is perturbed (Table~\ref{tab:corruption}). In the latter half of the reach, plane perturbations become stronger and close-range input dropout becomes more likely, reaching 25\,\% probability at closure. Dropout jointly blanks both image streams and zeroes the plane cue, preparing the policy for loss of instrument perception near the grasp. The \emph{w/o dropout} ablation disables this input dropout. Augmentation and random dropout are applied only during training.

\begin{table}[!ht]
\centering\scriptsize\setlength{\tabcolsep}{3pt}
\caption{Training-time perception augmentation. Ranges are uniform unless stated; $m$ denotes severity.}
\label{tab:corruption}
\begin{tabular}{p{0.3\columnwidth}p{0.62\columnwidth}}
\toprule
Component & Parameters\\
\midrule
Clean draw & $p=0.15$\\
Severity $m$ & $0.4$ ($p{=}0.40$), $0.8$ ($0.35$), $1.4$ ($0.10$), $1.0$ (rest)\\
Opening fill & $p=0.45$ (enclosed holes filled)\\
Dilate/erode & ellipse $k_x,k_y = 1+\lfloor\mathcal{U}\{1,2,3\}\cdot 2m\rfloor$; dilate $p{=}0.65$\\
Boundary wobble & SDF noise amplitude $\mathcal{U}(2,7)\,m$\,px, feature size 10--45\,px\\
False-positive blobs & $0..1{+}\lfloor 3m\rfloor$ blobs on the boundary, radius $\mathcal{U}(4,12)(0.5{+}m)$\\
Interior holes & $0..\lfloor 2m\rfloor$, radius $\mathcal{U}(3,10)(0.5{+}m)$\\
Half-plane cut & $p=0.03$\\
RGB leak fill & 85\,\% clutter (4--14 shapes, uneven light), 15\,\% plain; blur $k\le9$, noise $\sigma\in[2,8]$\\
Depth leak fill & random level $[0.05,1]$ + low-frequency field + 0--6 flat objects\\
Depth warp & $(\eta-0.5)\mathcal{U}(0.001,0.01)$ in normalized depth; $\eta\in[0,1]$ is smooth noise\\
Speckle & $\sigma=\mathcal{U}(0.002,0.012)(z/0.3)^2$\\
Quantization & step $\mathcal{U}(0.0008,0.0035)(z/0.3)^2$\\
Edge artifacts & Sobel band, 20--60\,\% flying pixels, 20--70\,\% dropout\\
Occlusion shadow & one-sided kernel half-width 2--7\,px, drop $p\in[0.4,0.9]$\\
Invalid patches & 0--4 blobs, radius 4--25\,px\\
Plane normal & $\mathcal{N}(0,3^\circ)$; gross $10$--$45^\circ$ with $p=0.03$; ramped toward closure, with joint input dropout\\
\bottomrule
\end{tabular}
\end{table}

\FloatBarrier

\subsection{Deployment in simulation and on the real robot}
\label{app:deployment}
\paragraph{Policy inputs and plane fit.}
Both deployments back-project masked depth pixels into a point cloud and estimate its normal by total least squares over the retained points $\{P_i\}_{i\in\mathcal I}$:
\begin{equation}
n=\arg\min_{\|n\|=1}\sum_{i\in\mathcal I}\big((P_i-\bar P_{\mathcal I})^{\top}n\big)^2 ,
\label{eq:plane}
\end{equation}
where $\bar P_{\mathcal I}$ is their centroid. The normal is the smallest right-singular vector of the centered cloud, oriented toward the camera. In simulation, depth filtering and residual trimming suppress outliers before refitting. On the real robot, the metric crop below removes the instrument tip.

\paragraph{Simulation.}
The evaluator obtains RGB-D and the instrument mask from the renderer, builds the same policy observation, and executes predicted joint-target chunks through the simulated arm controller at 25\,Hz. It queries again after each chunk. A completed close requires bilateral finger contact and a stalled gripper carriage for 12 control ticks; the simulated operator then releases the instrument, followed by 90 simulation updates for the retention check. Outcome criteria are given in Appendix~\ref{app:simscoring}.

\paragraph{Arm.}
The arm is a Trossen WidowX~AI: six revolute joints plus a parallel two-finger gripper (40\,mm maximum finger travel, 100\,N maximum grip), 0.77\,m reach, 1.5\,kg payload, 4\,kg mass, Ethernet position control with 500\,Hz feedback and a manufacturer-specified repeatability of about 1\,mm.

\paragraph{Camera.}
The wrist camera is an Intel RealSense D405 stereo depth camera: 18\,mm baseline, global-shutter imagers, $87^\circ\times58^\circ$ depth field of view, 7--50\,cm operating range with a minimum depth of 7\,cm at $848\times480$, and a manufacturer-specified Z accuracy of $\pm2$\,\% of distance with RMS spatial noise $\le 1$\,\% and temporal noise $\le 0.5$\,\%. At the 35--50\,cm working distance of this task these bounds correspond to up to 7--10\,mm of absolute depth error and 3.5--5\,mm of per-frame noise, which together with the arm's repeatability set the floor of the execution and perception variability discussed in \S\ref{sec:discussion}.

\paragraph{Real perception and control.}
D405 RGB and aligned depth are cropped to the simulated camera's field of view. SAM~3 tracks the instrument in video mode from the text prompt ``metal rake''; the largest detection is locked at first sighting and detections overlapping the gripper are discarded.

For all retractor types, we exclude the uppermost 1\,cm of the visible instrument mask, measured along its estimated three-dimensional longitudinal axis using depth and camera intrinsics, rather than a fixed pixel distance. Here, uppermost denotes the end higher in the wrist-camera image. The cropped mask is used for both plane estimation and the masked RGB-D observations supplied to the policy, so the plane is fitted only to the remaining visible body. This consistent preprocessing reduces orientation bias from bent tips, hooks and rake heads across retractor geometries.

SAM~3 tracking runs asynchronously; each policy query uses the latest completed RGB-D--mask observation. On an RTX~5090, warmed-up requests average 0.087\,s for $\pi_0$ and 0.078\,s for SmolVLA, excluding camera acquisition and logging. ACT with input dropout infers a 50-action chunk in approximately 10\,ms. The controller executes all 50 targets with light exponential smoothing and a per-tick joint clamp before requesting a new chunk. A completed close is detected when the commanded opening is below 20\,mm and the fingers close firmly until their motion stops. The policy then stops and the arm holds while the surgeon releases the instrument. The experimental protocol is given in Appendix~\ref{app:real}.
\FloatBarrier

\section{Evaluation and experimental details}
\label{app:eval}

\subsection{Test environments and presentations}
\label{app:real}
\paragraph{Simulation.}
All models use the same frozen set of 50 held-out presentations in the environment of Appendix~\ref{app:simenv}. The rendered instrument mask replaces the real segmenter, while the arm, gripper and camera configuration match deployment. The simulated hand and hook occluders provide local clutter; the scene has no abdominal phantom, operator or full operating-room surround.

\paragraph{Real experiment.}
The professionally fabricated abdominal phantom is designed for surgical-department simulation, with tissue-like appearance and texture. The arm operates beside the draped phantom under operating-room lighting (Figure~\ref{fig:robots}). The surgeon establishes retraction and passively holds the instrument while the robot approaches, then releases after the robot closes. Two attending surgeons defined 25 presentations each (50 total), spanning practical retraction directions. Each was photographed from two fixed viewpoints. Before each policy trial, the surgeon re-created the presentation and its instrument position and orientation were checked against both reference views. Appendix~\ref{app:rollouts} illustrates a simulated demonstration and a real rollout. The full model and its no-plane ablation also use 50 presentations per instrument (25 per surgeon) for the unseen Army--Navy and Richardson retractors. Presentation IDs denote corresponding surgeon-defined poses across all three instruments and all models to support paired comparisons.

\subsection{Simulation scoring}
\label{app:simscoring}
A simulated trial is \textbf{grasped} when the gripper closes on the retractor and its vertical drop after release is less than 20\,mm. Drop is measured from the instrument body's vertical position immediately before the simulated operator releases it to its position after 90 simulation updates. Failure to close on the instrument within 18\,s, or a drop of 20\,mm or more, counts as a failure.

For grasped trials, we report post-release instrument-tip offset and axial twist relative to the presented pose as mean $\pm$ standard deviation and median. These describe the motion introduced by take-over. Let $R_{\mathrm{pres}}$ and $R_{\mathrm{post}}$ map instrument coordinates to world coordinates before the approach and after release. Axial twist is the absolute $z$-component of $\operatorname{rotvec}(R_{\mathrm{post}}^{\top}R_{\mathrm{pres}})$, expressed in degrees; the instrument's local $z$-axis follows its shaft.

\subsection{Real outcomes and failure definitions}\label{app:failures}
\paragraph{Outcomes.}
Each real trial receives one of three mutually exclusive outcome labels.
\begin{itemize}
\item \textbf{Held:} the retractor remains in the gripper after release, maintaining tissue displacement and preserving the established retraction and exposure without repositioning at handoff.
\item \textbf{Grasped only:} the retractor remains in the gripper after release, but tissue displacement causes loss of the established retraction or exposure at handoff.
\item \textbf{Missed:} the robot does not secure the retractor, or the robot contacts the phantom, hand or retractor before the grasp begins, as defined below.
\end{itemize}
The reported \emph{grasped} rate includes both held and grasped-only trials; \emph{held} is the stricter task outcome (\S\ref{sec:task}).

\paragraph{Failure types.}
Each missed trial receives one failure type, assigned from the first decisive event:
\begin{itemize}
\item \textbf{Interference:} any robot contact with the phantom, surgeon's hand or retractor before the grasp begins. Recorded subtypes are \emph{inside} interference (contact with the instrument inside the gripper opening, including excessive insertion, contact with a bent instrument section at an incorrect angle, or a long instrument striking the pads, inner gripper structures or wrist camera), \emph{outside} interference (contact outside the gripper opening), \emph{hand}, and \emph{phantom}. Interference is always counted as a miss, regardless of any subsequent grasp.
\item \textbf{Lateral miss:} closure, or the run being stopped, with the shaft outside the pad opening and no contact.
\item \textbf{Short reach:} closure with the shaft inside the pad opening's projection along the approach axis, but not reached.
\end{itemize}

\paragraph{Observed mechanisms.}
Misaligned or off-center closure can push or rotate the shaft as the pads make contact, changing the established exposure even when the instrument is acquired. This is recorded as grasped only when acquisition succeeds but retraction is not preserved; any robot contact before the grasp begins instead counts as interference under the rule above.

Table~\ref{tab:interference} breaks down interference by contact location. The annotations group excessive insertion and instrument-length contact together. On the Army--Navy, trial 43 involves the bent section contacting the gripper at an incorrect angle. Without smoothing, 13 of 23 misses are lateral, including abrupt terminal motions that carry the gripper past the shaft before closure. On the training instrument, no inside-interference misses were observed with the full model, compared with three without dropout (unadjusted exact McNemar $p=0.25$). This pattern is consistent with the intended role of dropout under close-range perception ambiguity.

\begin{table}[!ht]
\centering\footnotesize\setlength{\tabcolsep}{3pt}
\caption{Interference counts by contact location, with 50 trials per condition. Each interference miss receives one subtype.}
\label{tab:interference}
\begin{tabular}{@{}lrrrrr@{}}
\toprule
Model & Inside & Outside & Hand & Phantom & Total\\
\midrule
\multicolumn{6}{@{}l}{Training retractor}\\
\method{} & 0 & 0 & 1 & 0 & 1\\
w/o dropout & 3 & 0 & 0 & 0 & 3\\
w/o recovery & 1 & 1 & 0 & 1 & 3\\
half data & 6 & 3 & 0 & 0 & 9\\
w/o plane & 2 & 1 & 0 & 0 & 3\\
w/o mask & 18 & 9 & 2 & 0 & 29\\
w/o smoothing & 0 & 3 & 1 & 1 & 5\\
w/o standoff & 5 & 3 & 4 & 2 & 14\\
SmolVLA & 2 & 3 & 2 & 0 & 7\\
ACT & 6 & 0 & 0 & 0 & 6\\
\midrule
\multicolumn{6}{@{}l}{Army--Navy}\\
\method{} & 2 & 0 & 0 & 0 & 2\\
w/o plane & 5 & 0 & 3 & 1 & 9\\
\midrule
\multicolumn{6}{@{}l}{Richardson}\\
\method{} & 3 & 1 & 0 & 1 & 5\\
w/o plane & 7 & 0 & 0 & 1 & 8\\
\bottomrule
\end{tabular}
\end{table}
\FloatBarrier

\begin{table*}[t]
\centering\small\setlength{\tabcolsep}{7pt}
\caption{Real-robot timing at default playback speed on grasped trials: mean $\pm$ SD [median]. Mean component latencies on an RTX~5090 are 86.6\,ms for $\pi_0$ requests, 77.6\,ms for SmolVLA requests, 152\,ms for SAM~3 first-frame detection, and 160\,ms for asynchronous tracking (median 98\,ms). ACT inference takes approximately 10\,ms per 50-action chunk.}
\label{tab:timing}
\begin{tabular}{@{}lcc@{}}
\toprule
Model & Time to grasp (s) & Policy queries (count)\\
\midrule
\method{} & 10.08 $\pm$ 1.69 [10.10] & 3.82 $\pm$ 0.49 [4]\\[3pt]
\method{}, Army--Navy & 12.41 $\pm$ 2.59 [12.30] & 3.90 $\pm$ 0.71 [4]\\[3pt]
\method{}, Richardson & 8.66 $\pm$ 1.10 [9.09] & 3.07 $\pm$ 0.26 [3]\\[3pt]
w/o dropout & 9.59 $\pm$ 1.62 [9.60] & 3.75 $\pm$ 0.49 [4]\\[3pt]
w/o recovery & 9.00 $\pm$ 1.69 [8.22] & 3.56 $\pm$ 0.55 [4]\\[3pt]
half data & 10.42 $\pm$ 1.93 [10.56] & 3.79 $\pm$ 0.56 [4]\\[3pt]
w/o plane normal & 9.70 $\pm$ 1.79 [9.56] & 3.76 $\pm$ 0.60 [4]\\[3pt]
w/o plane normal, Army--Navy & 11.70 $\pm$ 3.80 [10.15] & 3.82 $\pm$ 0.85 [4]\\[3pt]
w/o plane normal, Richardson & 10.15 $\pm$ 2.94 [9.35] & 3.43 $\pm$ 0.69 [3]\\[3pt]
w/o mask & 17.11 $\pm$ 4.79 [15.97] & 5.18 $\pm$ 0.87 [5]\\[3pt]
w/o smoothing & 9.14 $\pm$ 1.82 [9.42] & 3.52 $\pm$ 0.58 [4]\\[3pt]
w/o standoff & 8.52 $\pm$ 1.41 [8.31] & 3.21 $\pm$ 0.41 [3]\\[3pt]
SmolVLA & 10.25 $\pm$ 2.13 [9.88] & 3.92 $\pm$ 0.49 [4]\\[3pt]
ACT & 8.81 $\pm$ 1.39 [8.85] & 3.62 $\pm$ 0.58 [4]\\[3pt]
\bottomrule
\end{tabular}
\end{table*}

\paragraph{Trial conduct and annotation.}
Each trial permits one closure attempt. Motion stops at gripper closure or premature contact. Operators stop a trial if the approach stalls, moves away from the retractor, or the gripper closes without the retractor inside it; these trials count as missed. A separate operator runs the system and records the trial outcome. 
\section{Additional results}
\label{app:additional}

\subsection{Other policy architectures}
\label{app:arch}
Table~\ref{tab:arch} reports additional policy architectures using the same transfer framework. SmolVLA grasps 80\,\% in simulation and 74\,\% on the real robot, and holds 54\,\%. Its 13 real misses comprise seven interference and six lateral misses. ACT \citep{zhao2023act} grasps 52\,\% in simulation and 48\,\% on the real robot, with 16\,\% real held success: eight held, 16 grasped only, and 26 missed (17 lateral, six inside interference and three short reaches). Both real ACT rates are lower than the full model's (exact McNemar $p<0.001$ for each, unadjusted; Appendix~\ref{app:paired}). SmolVLA and ACT show descriptive simulation-to-real grasp-rate decreases of six and four percentage points, respectively, alongside lower simulated task success than $\pi_0$.

\begin{table}[!ht]
\centering\footnotesize\setlength{\tabcolsep}{3pt}
\caption{Additional policy architectures, with 50 trials per domain. Offset and twist over grasped simulated trials (Appendix~\ref{app:simscoring}) are mean $\pm$ SD, with medians beneath in brackets.}
\label{tab:arch}
\begin{tabular}{@{}lcc@{}}
\toprule
 & SmolVLA & ACT\\
\midrule
Sim. grasped & 80\,\% & 52\,\%\\
Offset (mm) & \shortstack{20.8 $\pm$ 17.2\\{[16.2]}} & \shortstack{12.3 $\pm$ 10.3\\{[9.8]}}\\
Twist ($^\circ$) & \shortstack{6.2 $\pm$ 10.0\\{[3.2]}} & \shortstack{6.4 $\pm$ 11.5\\{[3.9]}}\\
Real grasped & 74\,\% & 48\,\%\\
Real held & 54\,\% & 16\,\%\\
\bottomrule
\end{tabular}
\end{table}

\subsection{Paired outcome comparisons}
\label{app:paired}
We compare the full model with each ablation on 50 corresponding presentations, paired by ID. The primary outcome is \emph{held versus not held}, where not held includes both misses and grasped-only trials. The secondary outcome is \emph{grasped versus missed}, where grasped includes both held and grasped-only trials. Each uses an exact McNemar test, which compares presentations where only one model succeeds. Holm correction accounts for the seven ablation comparisons, separately for each outcome. Other paired comparisons report unadjusted $p$ values.

\begin{table}[!ht]
\centering\footnotesize\setlength{\tabcolsep}{3pt}
\caption{Paired held success: full model minus each ablation, in percentage points (50 presentations). Brackets show unadjusted 95\,\% paired bootstrap intervals; exact McNemar $p$ values are Holm-adjusted across seven comparisons.}
\label{tab:paired}
\begin{tabular}{@{}lcc@{}}
\toprule
Ablation & \shortstack{Held difference\\{[95\,\% CI], pp}} & $p_{\rm Holm}$\\
\midrule
w/o dropout & $+10\;[-8, 28]$ & 0.383\\
w/o recovery & $+44\;[26, 60]$ & $<0.001$\\
half data & $+60\;[42, 76]$ & $<0.001$\\
w/o plane & $+20\;[2, 38]$ & 0.128\\
w/o mask & $+66\;[52, 80]$ & $<0.001$\\
w/o smoothing & $+36\;[16, 56]$ & 0.006\\
w/o standoff & $+42\;[24, 60]$ & $<0.001$\\
\bottomrule
\end{tabular}
\end{table}

The full model has higher held success than the no-recovery, half-data, no-mask, no-smoothing and no-standoff variants (all adjusted $p<0.01$; Table~\ref{tab:paired}). For the secondary grasped outcome, the full model outperforms the no-mask, half-data and no-smoothing variants after correction.

For the full model, six corresponding presentations are not held on any of the three retractors; one of these is missed on all three. Across the eight main models, one presentation is missed by six; no presentation is missed by all eight. These shared failures suggest that photo-guided replication preserved challenging pose characteristics across instruments. The protocol deliberately samples a broad pose range to assess robustness across presentation configurations.

\subsection{Real-robot time to grasp}
\label{app:time}
Table~\ref{tab:timing} reports time to grasp and query counts for grasped trials at default demonstration playback speed. Time to grasp is wall-clock time from the first mask pass until the gripper stops closing. Control-tick timing quantifies inference overhead. SAM~3 tracking is asynchronous. For the full model, query overhead is about 0.89\,s of the 10.08\,s mean grasping time (8.8\,\%); execution and observation-related holds dominate. Playback speed can be increased without retraining, but hardware stability limits usable speeds: faster execution can increase control-induced drift with an unchanged policy.

\end{document}